\documentclass[letterpaper, 10 pt, conference]{ieeeconf}  

\IEEEoverridecommandlockouts                              

\usepackage{graphics} 
\usepackage{epsfig} 
\usepackage{amsmath} 
\usepackage{amssymb}  
\usepackage{xcolor}

\usepackage[space]{cite}

 \usepackage{color}

 \usepackage{subfiles}

\title{\LARGE \bf
Going In Blind: Object Motion Classification using Distributed Tactile Sensing for Safe Reaching in Clutter
}

\author{Rachel Thomasson$^{1}$, Etienne Roberge$^2$, Mark R. Cutkosky$^1$, and Jean-Philippe Roberge$^2$
\thanks{*This work was supported by Beijing Institute of Collaborative Innovation (BICI) and by Toyota Research Institute (TRI).}
\thanks{$^{1}$The authors are with the Mech. Eng. Dept., Stanford University,         Stanford, CA 94305, USA {\tt\small rthom@stanford.edu, cutkosky@stanford.edu}}%
\thanks{$^{2}$ The authors are with the Command and Robotics Laboratory, École de Technologie Supérieure, Montreal, Quebec, H3C1K3, Canada {\tt\small etienne.roberge.1@ens.etsmtl.ca, Jean-Philippe.Roberge@etsmtl.ca}}%
\thanks{\textcopyright 2022 IEEE. Personal use of this material is permitted. Permission
from IEEE must be obtained for all other uses, in any current or future
media, including reprinting/republishing this material for advertising or promotional purposes, creating new collective works, for resale or redistribution
to servers or lists, or reuse of any copyrighted component of this work in
other works.}%
}

\begin{document}

\maketitle
\thispagestyle{empty}
\pagestyle{empty}

\begin{abstract}

Robotic manipulators navigating cluttered shelves or cabinets may find it challenging to avoid contact with obstacles. Indeed, rearranging obstacles may be necessary to access a target. Rather than planning explicit motions that place obstacles into a desired pose, we suggest allowing incidental contacts to rearrange obstacles
while monitoring contacts for safety. 
Bypassing object identification, we present a method for categorizing object motions from tactile data collected from incidental contacts with a capacitive tactile skin on an Allegro Hand. We formalize tactile cues associated with categories of object motion, demonstrating that they can determine with $>90$\% accuracy whether an object is movable and whether a
contact is causing the object to slide stably (safe contact) or tip (unsafe).  

\end{abstract}

\section{Introduction}

When operating in human environments, robotic manipulators will encounter scenarios where obstacle avoidance is impractical. For instance, consider the task of retrieving an item from the back of a kitchen cabinet. It is likely that a collision-free path either does not exist or is intractable to visually identify due to occlusions. However, many of the obstacles in the scene will be movable and can therefore be pushed aside to create a path to the target. In situations such as these, humans are comfortable reaching towards the target and allowing, but not explicitly planning for, incidental contacts which rearrange the scene as necessary for task completion. They are able to sense and adapt to these contacts in order to maintain safety.


\begin{figure}[thpb]
  \centering
  \vspace{10pt}
  \includegraphics[width=0.9\linewidth]{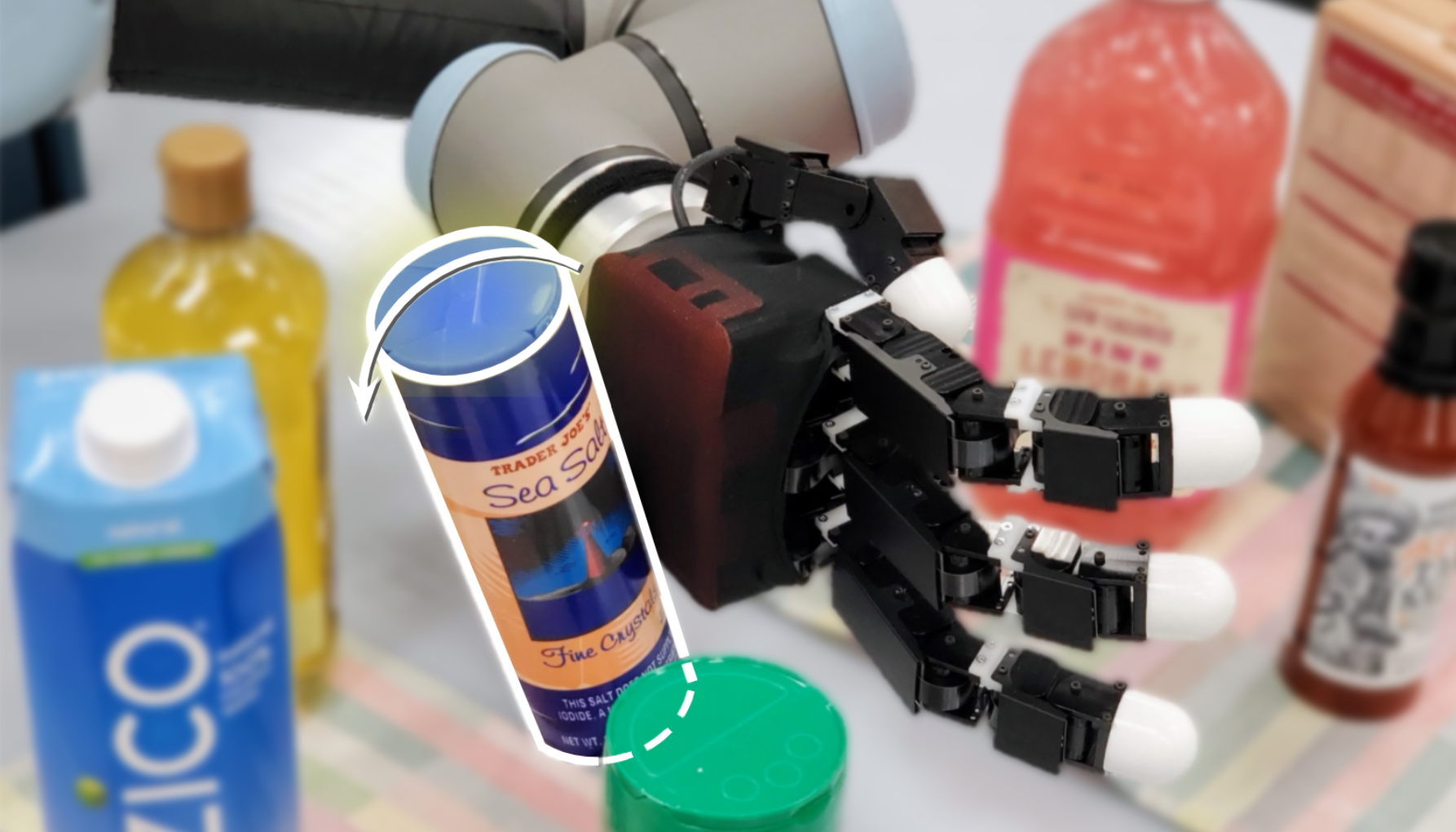}\vspace{-0.1in}
  \caption{When reaching through clutter, especially if the identity and mobility of objects are unknown, tactile sensors on surfaces such as the back of the hand can be used to prevent unsafe contact events, such as toppling an object.  \vspace{-0.15in}}
  \label{fig:motivation}
  \vspace{-5pt}
\end{figure}

This work is aimed towards enabling a robotic manipulator to use incidental contacts during reaching for safe, passive object rearrangement.  We do not focus on identifying the objects in contact and our goal is not to plan motions that push objects into a desired pose. Rather, we allow unplanned contacts to occur and aim to sense them and evaluate their safety. To minimize the need to reason about specific object properties, we propose that the safety of a contact can be inferred from the object motion it induces. The robot should be able to distinguish between contacts that stably reposition objects without knocking them out of their stable poses
(e.g. sliding an object across the surface of a shelf) and those that may be destructive (e.g. toppling an object or continuing to push against an object that is fixed in the world). As cluttered scenes present a challenge for vision systems due to occlusions and light contact events may produce little visual change, we rely on distributed tactile sensing. Whereas most tactile sensors developed for robotics applications focus on the fingertips and nominal grasping surfaces of manipulators, incidental contacts may occur anywhere on the hand and indeed may occur most often on the back of the hand when reaching toward a target. Therefore, we present a tactile skin design that covers not only the inner grasping surfaces, but also the back and sides of the 16 degree of freedom Allegro Hand. We then explore techniques for locally classifying object motion at the contacts. 

The contributions of this paper are:
\begin{itemize}
    \item an approach for reaching in clutter that uses tactile data to identify broad classes of object behavior and tactile cues that do not require identifying objects or object classes to predict behavior;
    \item a high-coverage, flexible sensor network for the Allegro Hand and its use to collect data during incidental contacts;
    \item demonstration of tactile cues for distinguishing immovable and movable objects, and further differentiating sliding (safe) and tipping (unsafe) motion.
\end{itemize}

\begin{figure*}[bht]
\vspace{5pt}
\centering
  \includegraphics[width=0.95
 \textwidth]{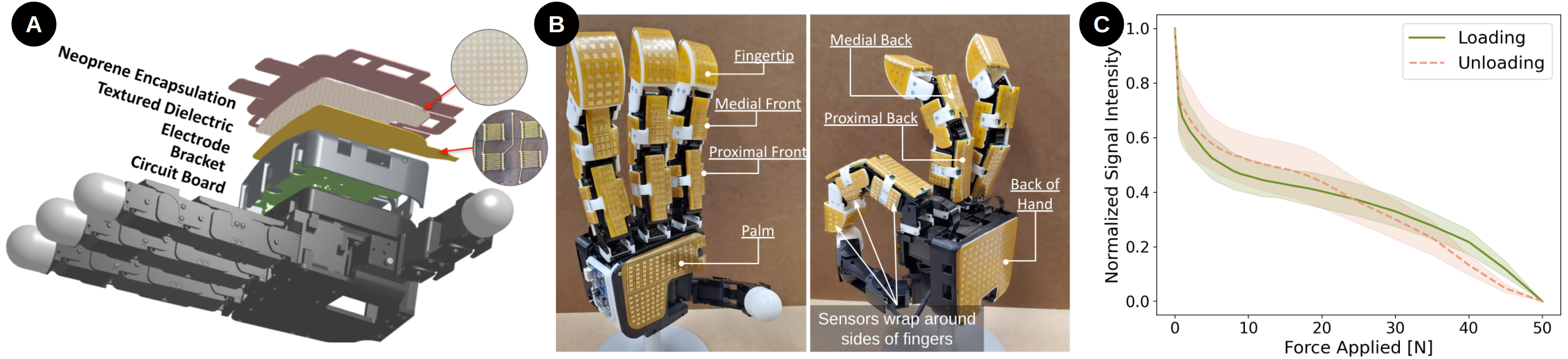}\vspace{-0.1in}
  \caption{(\textbf{A}) The layers that comprise a networked sensor, shown for the back of the hand. (\textbf{B}) Sensorized Allegro Hand with electrodes exposed for clarity. (\textbf{C}) Signal change as force is applied to the back of hand sensor. Solid and dashed lines show the mean signal of 10 taxels during loading and unloading, respectively, and shaded regions show the standard deviation. \vspace{-0.1in}
}
\label{fig:sensor-design}
\vspace{-5pt}
\end{figure*}

\subsection{Related Work}
\subsubsection*{Tactile Sensors and Sensory Skins}
There has been extensive work on developing tactile sensors for robotic end-effectors, mainly aimed at providing information for grasping and manipulation. Yamaguchi et al.~\cite{yamaguchi2019recent} provide a review of recent developments in tactile sensing technology for robotic manipulation. For the most part, these sensors cover nominal grasping surfaces, most commonly the fingertips \cite{wettels2014multimodal, maslyczyk2017highly, khamis2018papillarray,donlon2018gelslim, padmanabha2020omnitact, gomes2020geltip, lambeta2020digit}, but occasionally also the proximal inner surfaces and palm (e.g.~\cite{heyneman2016slip, funabashi2022multi}).
A few sensors have been developed specifically for, or integrated with, the commercial Allegro Hand, a 16 degree of freedom (DOF) robotic gripper \cite{funabashi2022multi, romero2020soft, sommer2016multi}, but these do not cover the backside of the hand. Some tactile skins that can potentially cover large areas of robotic arms (e.g.~\cite{gruebele2021stretchable, cirillo2015conformable, goncalves2021punyo})
have also been developed but are not specialized for a multi-fingered hand such as the Allegro Hand.

While optical tactile sensors offer remarkable resolution and sensitivity for manipulation~\cite{donlon2018gelslim, padmanabha2020omnitact, gomes2020geltip, lambeta2020digit}, capacitive sensors are easier to distribute across the different surfaces of a robotic hand and offer adequate resolution. The sensor presented here is adapted from a design by Ruth et al.~\cite{ruth2021flexible}, and is placed on the front, back, and sides of an Allegro Hand (Fig.~\ref{fig:sensor-design}).

\subsubsection*{Reaching Through Obstacles}
Previous studies have addressed challenges of manipulation in clutter by introducing a rearrangement subtask, where objects are repositioned using prehensile manipulations \cite{srivastava2014combined, murali20206}, non-prehensile pushes \cite{dogar2011framework, yuan2018rearrangement, hasan2020human, huang2020mechanical}, or a combination of both \cite{lee2021tree}. Zhong et al. ~\cite{zhong2022soft} track contacts along multiple objects during rummaging with soft tactile sensors. Moll et al.~\cite{moll2017randomized} propose to avoid reasoning about specific interactions and note that, in many cases, controlling the final pose of every object in the scene is unnecessary. They present an open-loop method which allows pushing movable objects while avoiding collisions with fixed obstacles, assuming that the mobility of each object is known \emph{a priori}. In this work, we use tactile sensing to detect object mobility, which is initially unknown and may change over time. Jain et al.~\cite{jain2013reaching} developed an MPC controller which enables a robotic manipulator with whole-arm tactile sensing to reach through obstacles while keeping contact forces below a threshold. We propose that using object motion as an indication of safety, rather than force, will allow sliding aside objects using forces which could topple objects in other scenarios. 

\subsubsection*{Object Mobility Classification} 
The benefit of tactile sensing in detecting object motion within a grasp, most notably the onset of slip, has been well discussed in literature \cite{yamaguchi2019recent}. Recent work from Ma et al.~\cite{ma2021extrinsic} uses vision-based distributed tactile sensing to estimate constraints to motion of a grasped object in contact with the environment. Using a novel sensor which provides both visual and touch information, Razaei et al.~\cite{rezaei2021learning} predict the final resting pose of an object that contacts the sensor after being released from an unstable configuration. There has been comparatively little work using tactile data from incidental contact on a robotic manipulator to classify object motion. Bhattacharjee et al.~\cite{bhattacharjee2014inferring, bhattacharjee2012haptic} use whole-arm tactile sensing to classify objects as Rigid-Fixed, Rigid-Movable, Soft-Fixed, and Soft-Movable through various data-driven approaches. Here, we expand on this work to differentiate between safe object motion (i.e. sliding) and potentially dangerous object motion (i.e. tipping) and propose a classification method which relies on designed tactile cues rather than a data-driven approach.  

\section{Distributed Tactile Sensor Design}
\label{sensor-design}
We now consider the design requirements for sensory skins intended to provide information about incidental contacts, focusing on classifying object motions, and present a new distributed tactile sensor network for the Allegro Hand. The skin used in this work is a modified version of the fringe field mutual capacitance sensor presented in \cite{ruth2021flexible}. The elements comprising each sensor in the network are outlined in Fig.~\ref{fig:sensor-design}A. A flexible PCB contains 3x3\,mm taxels with an interdigitated design arranged in a matrix. The electrodes are connected to the capacitance-to-digital converter of a Cypress PsoC 4200 microcontroller, which translates capacitance to signal counts. Pressure is transformed to a change in mutual capacitance through the addition of a textured dielectric which, when compressed, increases the effective dielectric constant as air is displaced. The textured dielectric is molded from urethane (SmoothOn Vytaflex-30) and features cylindrical pillars (0.8\,mm diameter, 0.8\,mm height). The dielectric is fixed to the electrode using a 1.6\,mm thick adhesive-backed silicone encapsulation layer. 

Contact localization is essential for determining how the robot should move in response to a contact. Especially when covering large areas, there is a trade-off between spatial resolution and sampling rate. In this work, we present experiments conducted with a back of hand sensor, 
which contains four taxels per $\text{cm}^2$ and has a sampling rate of 10\,Hz. Figure~\ref{fig:sensor-design}C shows the sensor's response to loading and unloading up to 50\,N as force is applied with a 12\,mm diameter probe. For a point contact (1.9\,mm diameter probe), the sensor can detect forces as light as 0.06\,N.

\subsection{Design Considerations for Interaction in Cluttered Spaces}
\subsubsection{Coverage}
Although contacts during grasping and manipulation may be constrained to specific areas of the hand, contacts during reaching often occur on the backside of the hand, especially if reaching with fingers slightly flexed. To avoid undetected contacts, the skin should ideally cover the entire hand. The coverage achieved by our sensor network is displayed in Fig. \ref{fig:sensor-design}B. 

\subsubsection{Curvature}
\begin{figure}[thb]
\vspace{5pt}
  \centering
  \includegraphics[width=\linewidth]{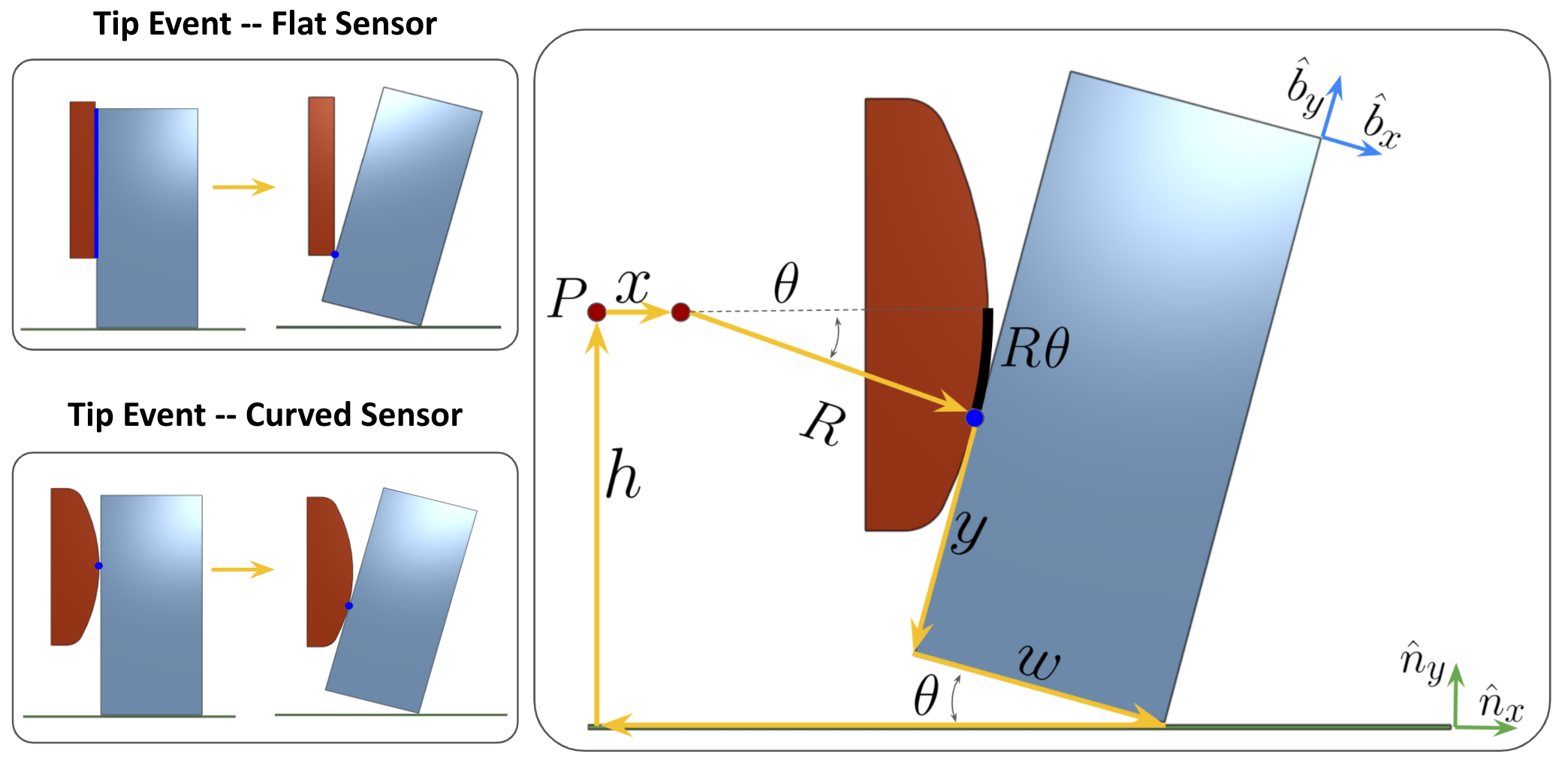}
  \caption{Curved surfaces are beneficial for sensing tipping. A schematic illustrates the vector loop in eq.~\ref{eqn:loop}.}
  \label{fig:curved-sensor}
  \vspace{-12pt}
\end{figure}

While flat tactile sensors are easier to manufacture, curvature is beneficial for capturing and tracking incidental contact events. As shown in Fig.~\ref{fig:curved-sensor}, flat sensors provide large contact areas when aligned with a surface, but are prone to quickly losing contact when there is any misalignment, such as the onset of tip. Adding curvature to the sensor in the direction of gravity leads to contacts which travel down the surface of the sensor as an object tips. 

We develop a 2D kinematic model for contact point motion as a curved sensor tips an object. This will later be used to determine appropriate tactile cues for tip events. For a sensor with radius of curvature $R$ contacting a vertically uniform object of width $w$, one can determine the tip angle, $\theta$, of the object by solving the vector loop equation set up in Fig.~\ref{fig:curved-sensor}. We define two right-handed orthonormal bases: $\hat{n}_{xyz}$ and $\hat{b}_{xyz}$, fixed to the world and object, respectively. Unit vector $\hat{n}_x$ points along the ground and $\hat{n}_y$ points vertically upward and, initially, $\hat{b}_x = \hat{n}_x$ and $\hat{b}_y = \hat{n}_y$. The sensor contacts the object at height $h$ and moves along $\hat{n}_x$ a distance of $x$, causing the object to tip by $\theta$. We write a vector loop beginning and ending at $P$, the center of curvature of the sensor when it initially contacts the object, to obtain
\begin{equation}
    x \hat{n}_x + R \hat{b}_x - y \hat{b}_y + w \hat{b}_x - (R + w)\hat{n}_x = \Vec{0}. 
    \label{eqn:loop}
\end{equation}
Taking the dot product of eq.~\ref{eqn:loop} with $\hat{n}_x$ and $\hat{n}_y$ produces two scalar equations which can be solved for $\theta$. The distance we expect the contact to move in the gravity direction can then be calculated as $R \theta$. Smaller radii may require a larger tip angle before contact patch motion indicating tip is detectable but larger radii will lose contact with a tipping object faster. The sensor used in experiments features a gentle curvature with a radius of 372\,mm. 

\section{Technical Approach}
\subsection{Environment Assumptions}
\label{assumptions}
We consider a robot performing a planar reaching task in the presence of obstacles. Robot trajectories lie in a horizontal plane and consist of piecewise linear motions. Operating speeds are low enough that a quasi-static assumption applies (i.e. there is no dynamic evolution of the scene that is not associated with robot motion). We assume contacts occur on a sensorized surface of the hand. Objects may vary in geometry, but we assume they are convex, rigid, and feature a stable support surface which is initially in contact with the surface of the world. We note that non-convex objects can often be treated as locally convex if considering only their sliding or tipping motion. The mobility of objects is initially unknown.
They may be free to slide, constrained such that they are likely to tip, or immovable in certain directions. We consider any object which takes more than 20\,N to move to be immovable.

\subsection{Terminology}
We use the following terms to formulate expectations about how tactile signals evolve in response to various object behaviors: \textbf{Tactile Feature:} a quantity extracted from tactile data at an instant. Examples include shape, size, location, and intensity of contact patches. \textbf{Tactile Cue:} a measure of how tactile features evolve over time -- for instance, an increase in contact patch size. \textbf{Tactile Event:} the tactile response corresponding to an object behavior (tipping, sliding, or fixed contact events). A tactile event may correspond to one or more tactile cues. 

\subsection{Object Motion Categories}
\label{motion-categories}
In general, object motion from incidental contact is complex and depends on the object's geometric and mass properties. For determining the safety of a contact interaction, it may suffice to broadly understand the object's motion category rather than estimating its trajectory. We consider the following categories:
\begin{itemize}
    \item \textbf{Immovable} in the direction of applied force either because it is fixed in the environment, constrained, or considered too heavy to push. 
    \item \textbf{Sliding} across the working surface while remaining in its stable pose. 
    \item \textbf{Tipping} out of its stable pose. Excessive tip may lead to toppling, potentially causing damage or spills. 
\end{itemize}
We hypothesize that these broad motion categories can be inferred from tactile cues that are not highly object-dependent. Therefore, using tactile signals to categorize object behavior could allow robots to reason about the safety of their actions without needing to identify the object itself. There also exist objects without stable poses (e.g. spheres), which we do not explicitly consider here. In addition to sliding, these objects may also roll, which would produce similar tactile signals to sliding and is also considered locally safe. 

\begin{figure}[t!]
\vspace{5pt}
  \centering
  \includegraphics[width=1\linewidth]{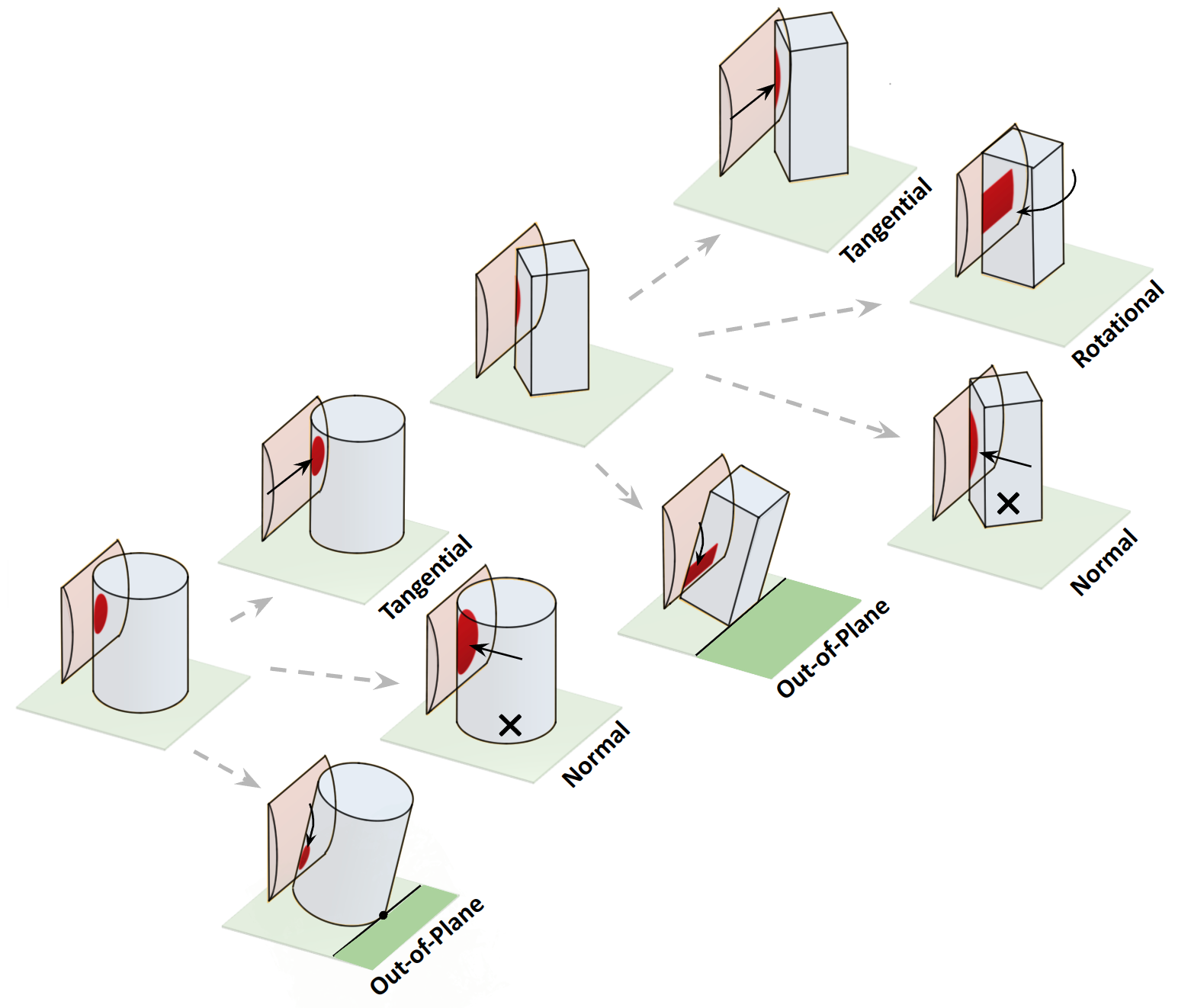}
  \caption{We can distinguish two planar DoFs for radially symmetric objects (e.g. cylinders) and three for radially asymmetric objects (e.g. boxes).  Out of plane motion (tip) provides another distinguishable DoF.}
  \label{fig:cp-ecpectations}
  \vspace{-10pt}
\end{figure}

Prior to choosing potential cues which indicate motion category, we note that radial symmetry of an object impacts the tactile signals we expect. In their stable poses, objects that are radially symmetric about the vector normal to the floor have two detectable degrees of freedom: motion in the directions normal and tangential to the sensor surface. Radially asymmetric objects have three detectable degrees of freedom in their stable poses, as rotation in the plane will affect the sensor signal, as illustrated in Fig.~\ref{fig:cp-ecpectations}.

\subsection{Tactile Feature Extraction}
\label{feature-extraction}

\begin{figure}[b!]
  \centering
  \includegraphics[width=1\linewidth]{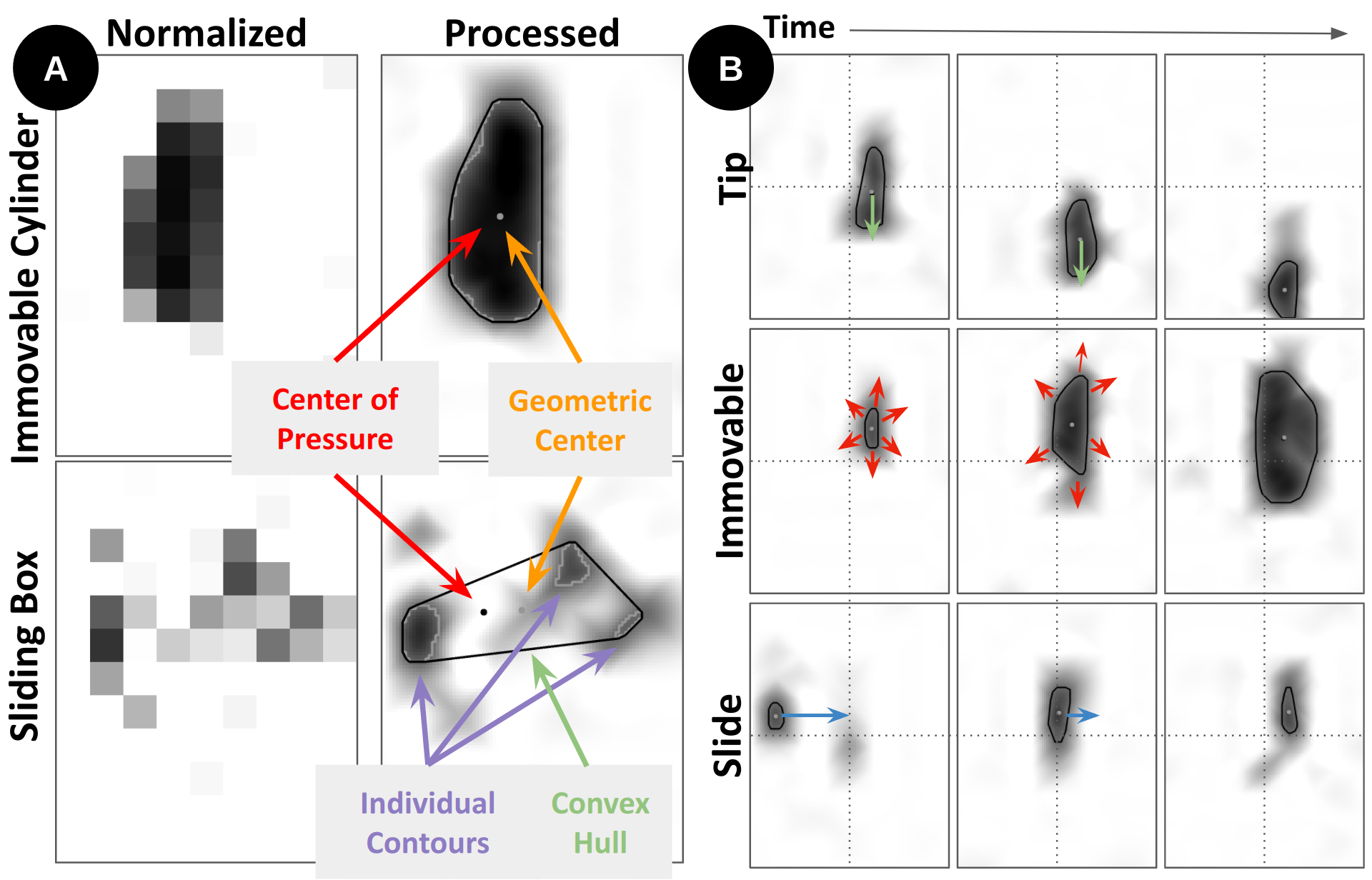}
  \caption{(\textbf{A}) Snapshot tactile images taken from contact with a fixed cylinder and sliding box are shown at two stages of post-processing. Features extracted from the processed images are labeled. (\textbf{B}) Sequences of tactile images in which tactile events can be identified. }
  \label{fig:feature-extraction}
\end{figure}

Our candidate tactile features are: (1) [x,y] location of the contact patch center of pressure, (2) contact patch area, and (3) intensity of the activated taxels within the contact patch. They are identified using the following processing steps. First, we represent the normalized data at each timestep as a greyscale image. Signal intensity is extracted at this step. We then use bicubic interpolation to obtain sub-taxel resolution. Next, thresholding is used to isolate the contact patch. The same threshold value was used across experiments. Image moments are used to calculate the center of pressure of the contact patch, $(C_x, C_y)$, using $C_x = \frac{M_{10}}{M_{00}}$ and $C_y = \frac{M_{01}}{M_{00}}$ where $M_{ij}$ is the $(i, j)^{th}$ image moment. Finally, a contour which surrounds the patch is found. To accommodate cases where contact with a single object results in multiple disjoint patches (due to sensor noise or light, uneven contact), we take the convex hull of all contours to combine them into a single patch before calculating area as illustrated in Fig.~\ref{fig:feature-extraction}A.

\subsection{Tactile Cue Extraction}
\label{tactile-cue-extraction}
Candidate tactile cues such as an increase in signal intensity or a contact patch motion may be straightforward to identify conceptually, but reliably deriving them from raw tactile data is more involved. While theoretically, one could consider a tactile cue to be a discrete change in a tactile feature between two points in time, we observe that it is more reliable to identify trends over sequences of tactile data. Indeed, due to real world factors such as sensor noise, uneven contact surfaces, and variations in friction between the objects and the table, discrete changes in the contact patch often instantaneously resemble tactile events that are not actually present. We perform a piecewise linear least squares fit on the evolution of extracted tactile features either over time or over the progression of a robot motion \cite{pwlf}. Two line segments are used to capture a change in object motion partway through the interaction (e.g. objects may initially slide then begin to tip). For each tactile feature, we consider the slopes and means over each line segment to be candidate tactile cues. Before performing the linear regression, the features are passed through the RANSAC algorithm to remove outliers caused by noise. 

\section{Experiments}
We now describe the process for collecting incidental contact data and discuss the selection process for tactile cues. We then present classification results for contacts with immovable, tipping, and sliding objects. 

\subsection{Incidental Contact Data Collection}
\subsubsection{Experimental Setup}
Incidental contact data were collected using the sensor presented in Sec. \ref{sensor-design} on the back of the Allegro Hand, which was mounted on a UR16e robotic arm and stationed above a flat table. The hand was placed in a configuration suitable for reaching, with the fingers extended and the palm orthogonal to the table surface, as in Fig.~\ref{fig:motivation}. Future work will apply the approach demonstrated here on the back of the hand to the entire hand sensor network.

\begin{figure}[t!]
  \centering
  \vspace{10pt}
  \includegraphics[width=0.8\linewidth]{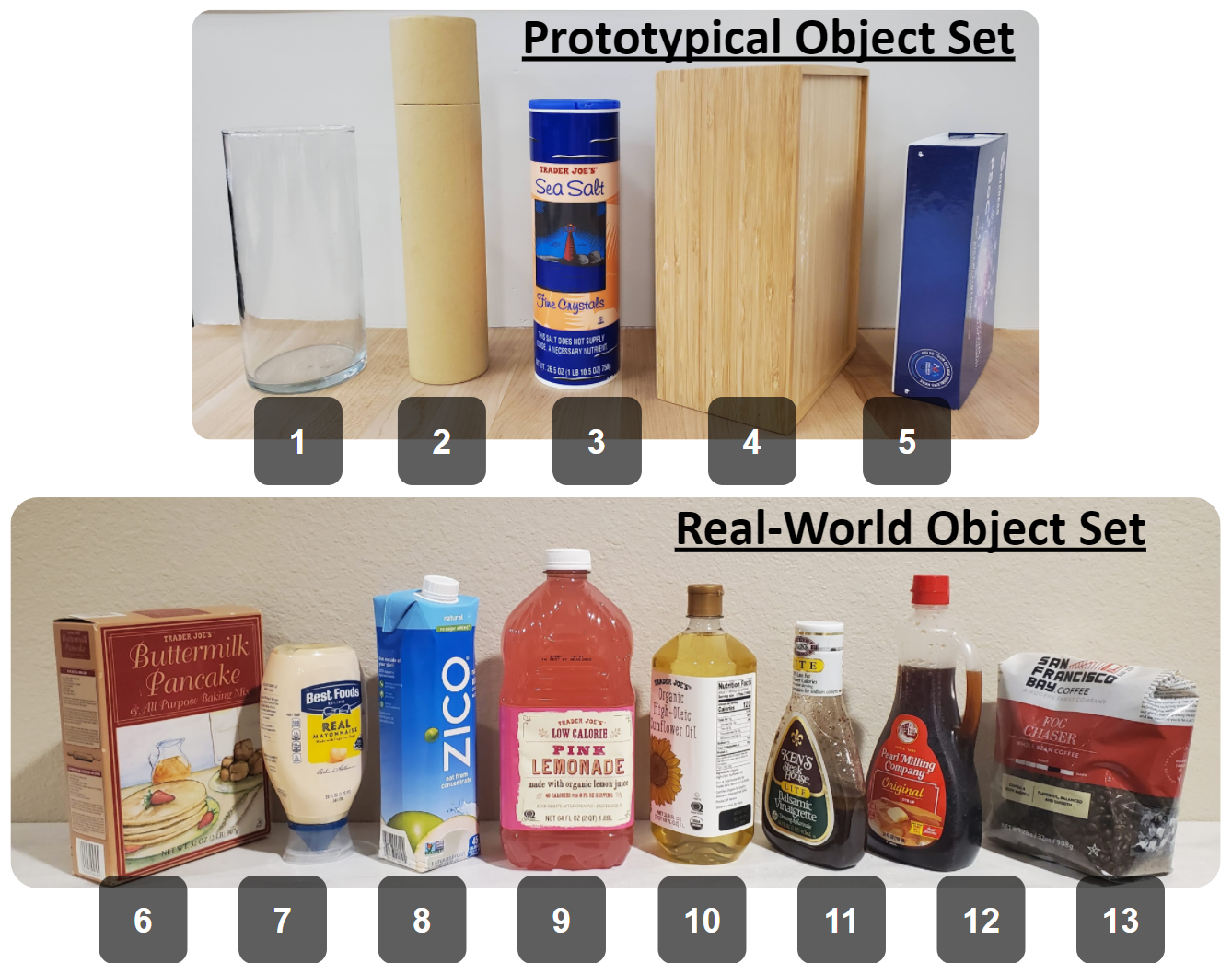}
  \caption{Two sets of objects used for experiments. Object dimensions and weights are listed in Table \ref{tab:objects}.}
  \label{fig:objects}
\end{figure}

\begin{table}[hbt]
    \centering
\begin{tabular}{ |l|c|c|c| } 
\hline
Object & Dimensions [mm] & Weight [g] \\
\hline
(1) Glass Vase & 90 x 190 & 390  \\ 
(2) Cardboard Cylinder & 61 x 270 & 170   \\ 
(3) Salt Shaker & 68 x 208 & 910 \\ 
(4) Wooden Box & 100 x 165 x 230 & 420 \\ 
(5) Cardboard Box & 43 x 133 x 180 & 260 \\ 
(6) Pancake Mix & 60 x 155 x 208 & 960 \\ 
(7) Mayonnaise & 70 x 90 x 180 & 610 \\
(8) Coconut Water & 74 x 80 x 230 & 1060  \\ 
(9) Lemonade & 80 x 110 x 270 & 2000 \\ 
(10) Sunflower Oil & 76 x 76 x 240 & 950  \\ 
(11) Balsamic Vinaigrette & 45 x 103 x 204 & 300 \\
(12) Syrup & 70 x 106 x 255 & 900 \\
(13) Coffee Grounds & 90 x 140 x 180 & 620 \\ 
\hline
\end{tabular}
    \caption{Object Properties}
    \label{tab:objects}
    \vspace{-15pt}
\end{table}

\subsubsection{Object Sets}
We collected data for contact with two sets of objects, shown in Fig.~\ref{fig:objects}. The first set contains prototypical objects which are approximately rigid and geometrically simple (2 boxes and 3 cylinders). We expect motion categories for these objects to be separable using the tactile cues discussed in Sec.~\ref{tactile-cue-extraction}. The second set consists of eight additional commonplace kitchen objects. These objects display a wider range of geometries than the prototypical object set which led to more varied object motions in response to contact. While our assumption was that objects are rigid, many of these are slightly deformable. One object featured concavities (lemonade) and three featured support surfaces that, while visually appearing flat, rested unevenly on the table leading to wobbly motion during sliding (pancake mix, coconut water, coffee beans). The purpose of the second object set is to investigate the robustness of our classification method to different object geometries and to identify and discuss failure cases.

\subsubsection{Data Collection}
Data were collected for objects individually. For each trial, an object was placed on the table in one of three conditions: (1) free-standing without constraints, (2) placed along the edge of a flat obstacle (a placemat) which promoted tipping, or (3) constrained against a wall. The arm was then commanded to move along pre-specified Cartesian paths which extended beyond the object, leading to a collision with the back of hand sensor. For each object, 20 trials were taken for each mobility condition -- half for which the tangential component of motion was zero and half for which it was equal to the normal component. Ground truth class labels were assigned by hand. The robot was commanded to stop if the force in the direction normal to the sensor surface exceeded 30\,N. This was monitored with the integrated F/T sensor at the UR16e wrist, but F/T data were not used for object motion classification. Between trials, the initial pose of the objects was manually shifted such that the object and sensor came into contact at different locations. To limit inertial effects, we command the robot to move at a speed of 10\,mm/s for all trials. 

\begin{figure}[b!]
  \centering
  \includegraphics[width=\linewidth]{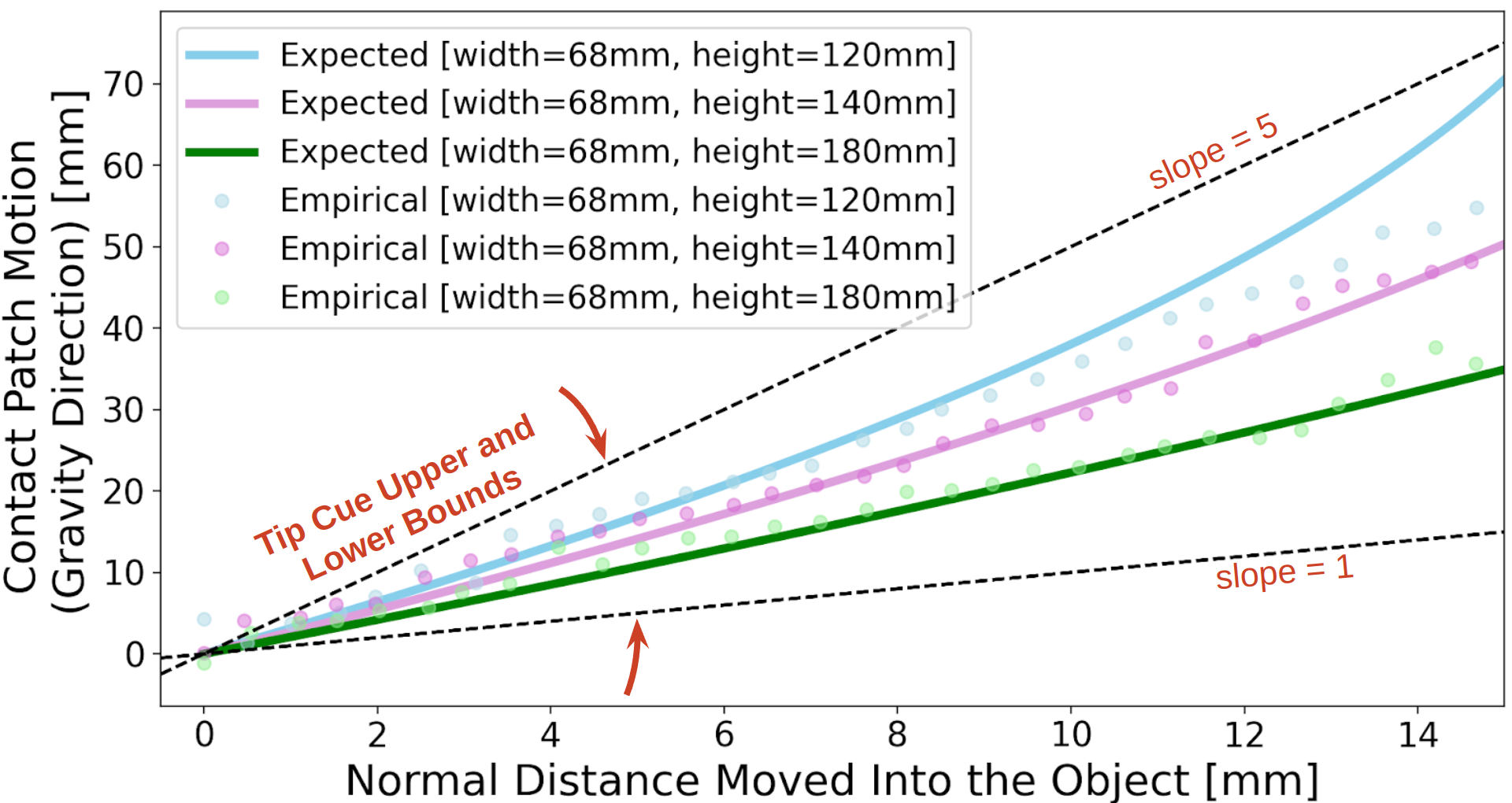}
  \caption{Empirical tip data plotted over expected contact patch motion during tip at three heights, calculated as $R\theta$ where $\theta$ is found by solving eq.~\ref{eqn:loop}.}
  \label{fig:tip-graph}
\end{figure}

\begin{figure*}[t!]
\centering
\vspace{10pt}
  \includegraphics[width=
 \textwidth]{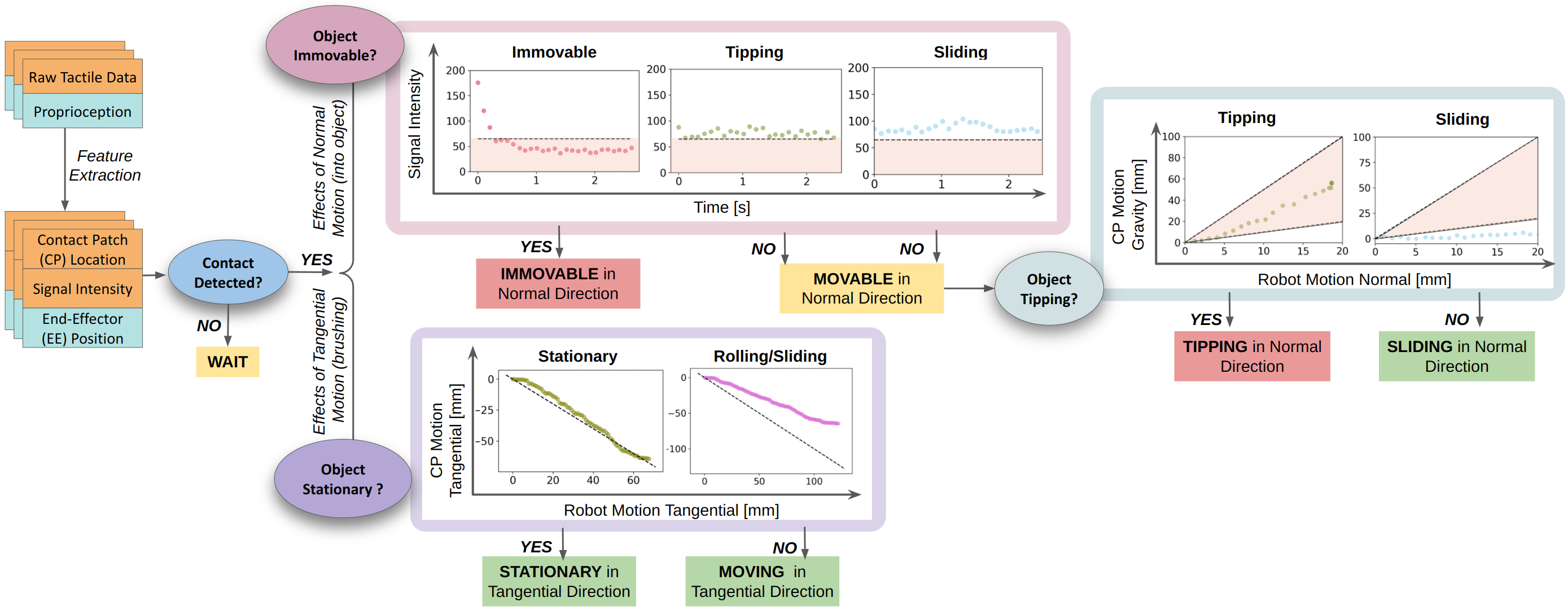}
  \caption{A flowchart summarizing the steps for determining object motion category from tactile and proprioception data. Plots show real experimental data along with the conditions that signals must meet for tactile events to be predicted. 
}
\label{fig:classifier}
\vspace{-12pt}
\end{figure*}

\subsection{Tactile Cue Selection}
Preliminary incidental contact data were used to identify which cues would be used for classification and to establish appropriate threshold values. To test the ability of eq.~\ref{eqn:loop} to predict tipping based on contact patch motion, we collected empirical results for motion of an object with known width being tipped by contact at three heights. Center of pressure locations calculated using the feature extraction method described in Sec.~\ref{feature-extraction} are plotted against analytical results in Fig.~\ref{fig:tip-graph}. To accommodate for a range of object widths, curvatures, and collision heights, we estimate that an object is tipping if the rate at which the contact patch moves in the direction of gravity as the robot moves into the object is between 1 and 5 (slopes shown in Fig.~\ref{fig:tip-graph}). 

To identify contact with an immovable object, we hypothesized that signal intensity and contact patch growth could both serve as cues. However, it was observed from preliminary data that contact patch growth was highly variable between objects and trials, largely due to orientation changes of radially asymmetric objects. On the other hand, signal intensity, calculated from normalized tactile images as the mean value of taxels at the center of the contact patch, was found to be a good separator of contact with immovable and movable objects. As the signal for a mutual capacitance pressure sensor decreases with increasing pressure, we identify fixed contact if the average signal intensity (over one segment of the piecewise linear fit for signal intensity versus time data)  is less than 65. Figure~\ref{fig:classifier} illustrates the use of these cues for object motion classification, depicting the thresholds defined and examples of real data for each category.

\begin{figure}[b!]
  \centering
  \includegraphics[width=1\linewidth]{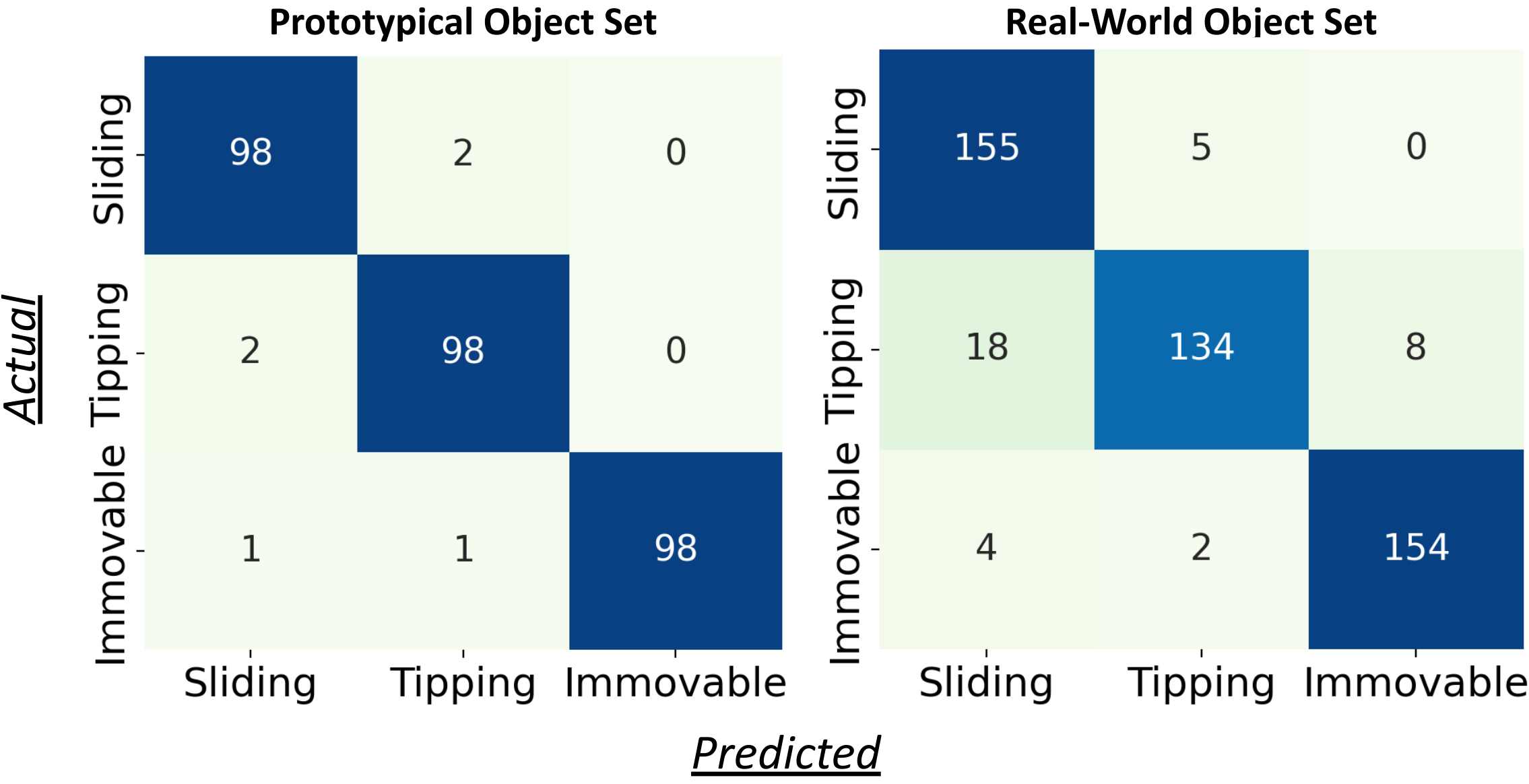}
  \caption{Confusion matrices containing the result of classifying incidental contact with immovable, tipping, and sliding objects. Numerical results indicate number of trials.}
  \label{fig:confusion-matrices}
\end{figure}

\subsection{Classification Results \& Discussion}
Figure \ref{fig:confusion-matrices} shows classification results for incidental contact with immovable, tipping, and sliding objects. For objects in the prototypical set, simple and object-independent cues achieve 98\% accuracy in separating motion classes. Though the misclassifications are few, we note an interesting failure case for box-like objects. In a few cases where a box was placed such that it was likely to tip about a corner rather than a flat edge, contact caused it to tip in a direction orthogonal to the normal force, which was undetectable by the sensor. 

Results for the more geometrically diverse real-world objects showed a combined accuracy of 92\%, where most confusion comes from mistaking tipping for sliding motion. 14 of the 18 misclassifications of tipping as sliding arose from two objects: coffee beans and lemonade. Deformable and granular coffee beans interacted with the sensor differently than rigid objects during tip events. During tip, the sensor would deform the object, causing a contact patch that often appeared nearly stationary rather than moving in the direction of gravity. When contact causing tip of the lemonade container occurred on areas featuring concavities, contacts did not travel smoothly down the length of the object, as we would expect with a convex object. Rather, the portion of the container that protruded the most would remain in contact with the sensor, causing the contact patch to travel upwards along the sensor rather than downwards. For difficult cases such as these, where rigidity and convexity assumptions do not hold, it may be necessary to identify the object prior to estimating motion. A complementary video is provided to demonstrate the use of this classification method.

\subsection{Analyzing Tangential Motion}
While the focus of this work is classifying object motion induced by applying normal force, we note that useful information can be extracted from tangential components of the data as well. By comparing the horizontal contact patch motion to the robot's tangential motion, one can estimate whether an object is stationary or moving as a result of the robot brushing against it. If horizontal contact patch motion is equal and opposite to the robot's motion, this indicates a stationary object. Otherwise, the brushing contact is causing the object to roll or slide. Figure \ref{fig:classifier} shows this distinction for data collected as the robot moves tangentially while in contact with a cylindrical object placed against a wall (fixed in the normal direction) both when the object is stationary and when it exhibits a combination of rolling and sliding.

\section{Conclusions and Future Work}
We present a method that uses incidental contact data from a tactile sensor on the back of a robotic hand to classify object motion. We show that our formulation of tactile cues is able to capture useful information about whether objects are movable or immovable and, if movable, whether contact results in safe sliding motion or potentially dangerous tipping. By using tactile data to understand the motions induced by incidental contact events, robotic manipulators reaching in clutter can rearrange objects without explicitly planning rearrangement motions or identifying the object in contact. 

Future work involves applying this classification method to the full sensor network and developing reactive control methods which modify a robot's trajectory when a dangerous object motion is detected. A set of exploratory probing techniques could be designed to determine whether an object classified as immovable is completely fixed, or constrained only to move in certain directions. Additionally, methods for analyzing normal and tangential motion components -- perhaps with shear as well as normal tactile data -- could be combined to estimate objects' trajectories in more detail for higher level planning.

\addtolength{\textheight}{-12cm}   


\section*{ACKNOWLEDGMENT}

The Beijing Institute of Collaborative Innovation (BICI) and Toyota Research Institute (TRI) provided funds to support this work. Added thanks to TRI for their help with the Allegro Hand hardware and setup. R.~Thomasson is additionally supported by the NSF Graduate Fellowship.
The authors thank Hojung Choi, Michael Lin, and Anthony Chen for their helpful discussions and insightful suggestions!


\bibliographystyle{unsrt}
\bibliography{citations}

\begin{thebibliography}{10}

\bibitem{yamaguchi2019recent}
Akihiko Yamaguchi and Christopher~G Atkeson.
\newblock Recent progress in tactile sensing and sensors for robotic
  manipulation: can we turn tactile sensing into vision?
\newblock {\em Advanced Robotics}, 33(14):661--673, 2019.

\bibitem{wettels2014multimodal}
Nicholas Wettels, Jeremy~A Fishel, and Gerald~E Loeb.
\newblock Multimodal tactile sensor.
\newblock In {\em The Human Hand as an Inspiration for Robot Hand Development},
  pages 405--429. Springer, 2014.

\bibitem{maslyczyk2017highly}
Alexis Maslyczyk, Jean-Philippe Roberge, Vincent Duchaine, et~al.
\newblock A highly sensitive multimodal capacitive tactile sensor.
\newblock In {\em 2017 IEEE International Conference on Robotics and Automation
  (ICRA)}, pages 407--412. IEEE, 2017.

\bibitem{khamis2018papillarray}
Heba Khamis, Raquel~Izquierdo Albero, Matteo Salerno, Ahmad~Shah Idil, Andrew
  Loizou, and Stephen~J Redmond.
\newblock Papillarray: An incipient slip sensor for dexterous robotic or
  prosthetic manipulation--design and prototype validation.
\newblock {\em Sensors and Actuators A: Physical}, 270:195--204, 2018.

\bibitem{donlon2018gelslim}
Elliott Donlon, Siyuan Dong, Melody Liu, Jianhua Li, Edward Adelson, and
  Alberto Rodriguez.
\newblock Gelslim: A high-resolution, compact, robust, and calibrated
  tactile-sensing finger.
\newblock In {\em 2018 IEEE/RSJ International Conference on Intelligent Robots
  and Systems (IROS)}, pages 1927--1934. IEEE, 2018.

\bibitem{padmanabha2020omnitact}
Akhil Padmanabha, Frederik Ebert, Stephen Tian, Roberto Calandra, Chelsea Finn,
  and Sergey Levine.
\newblock Omnitact: A multi-directional high-resolution touch sensor.
\newblock In {\em 2020 IEEE International Conference on Robotics and Automation
  (ICRA)}, pages 618--624. IEEE, 2020.

\bibitem{gomes2020geltip}
Daniel~Fernandes Gomes, Zhonglin Lin, and Shan Luo.
\newblock Geltip: A finger-shaped optical tactile sensor for robotic
  manipulation.
\newblock In {\em 2020 IEEE/RSJ International Conference on Intelligent Robots
  and Systems (IROS)}, pages 9903--9909. IEEE, 2020.

\bibitem{lambeta2020digit}
Mike Lambeta, Po-Wei Chou, Stephen Tian, Brian Yang, Benjamin Maloon,
  Victoria~Rose Most, Dave Stroud, Raymond Santos, Ahmad Byagowi, Gregg
  Kammerer, et~al.
\newblock Digit: A novel design for a low-cost compact high-resolution tactile
  sensor with application to in-hand manipulation.
\newblock {\em IEEE Robotics and Automation Letters}, 5(3):3838--3845, 2020.

\bibitem{heyneman2016slip}
Barrett Heyneman and Mark~R Cutkosky.
\newblock Slip classification for dynamic tactile array sensors.
\newblock {\em The International Journal of Robotics Research}, 35(4):404--421,
  2016.

\bibitem{funabashi2022multi}
Satoshi Funabashi, Tomoki Isobe, Fei Hongyi, Atsumu Hiramoto, Alexander
  Schmitz, Shigeki Sugano, and Tetsuya Ogata.
\newblock Multi-fingered in-hand manipulation with various object properties
  using graph convolutional networks and distributed tactile sensors.
\newblock {\em IEEE Robotics and Automation Letters}, 7(2):2102--2109, 2022.

\bibitem{romero2020soft}
Branden Romero, Filipe Veiga, and Edward Adelson.
\newblock Soft, round, high resolution tactile fingertip sensors for dexterous
  robotic manipulation.
\newblock In {\em 2020 IEEE International Conference on Robotics and Automation
  (ICRA)}, pages 4796--4802. IEEE, 2020.

\bibitem{sommer2016multi}
Nicolas Sommer and Aude Billard.
\newblock Multi-contact haptic exploration and grasping with tactile sensors.
\newblock {\em Robotics and autonomous systems}, 85:48--61, 2016.

\bibitem{gruebele2021stretchable}
Alexander~M Gruebele, Michael~A Lin, Dane Brouwer, Shenli Yuan, Andrew~C Zerbe,
  and Mark~R Cutkosky.
\newblock A stretchable tactile sleeve for reaching into cluttered spaces.
\newblock {\em IEEE Robotics and Automation Letters}, 6(3):5308--5315, 2021.

\bibitem{cirillo2015conformable}
Andrea Cirillo, Fanny Ficuciello, Ciro Natale, Salvatore Pirozzi, and Luigi
  Villani.
\newblock A conformable force/tactile skin for physical human--robot
  interaction.
\newblock {\em IEEE Robotics and Automation Letters}, 1(1):41--48, 2015.

\bibitem{goncalves2021punyo}
Aimee Goncalves, Naveen Kuppuswamy, Andrew Beaulieu, Avinash Uttamchandani,
  Katherine~M Tsui, and Alex Alspach.
\newblock Punyo-1: Soft tactile-sensing upper-body robot for large object
  manipulation and physical human interaction.
\newblock {\em arXiv preprint arXiv:2111.09354}, 2021.

\bibitem{ruth2021flexible}
Sara Rachel~Arussy Ruth, Vivian~Rachel Feig, Min-gu Kim, Yasser Khan,
  Jason~Khoi Phong, and Zhenan Bao.
\newblock Flexible fringe effect capacitive sensors with simultaneous
  high-performance contact and non-contact sensing capabilities.
\newblock {\em Small Structures}, 2(2):2000079, 2021.

\bibitem{srivastava2014combined}
Siddharth Srivastava, Eugene Fang, Lorenzo Riano, Rohan Chitnis, Stuart
  Russell, and Pieter Abbeel.
\newblock Combined task and motion planning through an extensible
  planner-independent interface layer.
\newblock In {\em 2014 IEEE international conference on robotics and automation
  (ICRA)}, pages 639--646. IEEE, 2014.

\bibitem{murali20206}
Adithyavairavan Murali, Arsalan Mousavian, Clemens Eppner, Chris Paxton, and
  Dieter Fox.
\newblock 6-dof grasping for target-driven object manipulation in clutter.
\newblock In {\em 2020 IEEE International Conference on Robotics and Automation
  (ICRA)}, pages 6232--6238. IEEE, 2020.

\bibitem{dogar2011framework}
Mehmet Dogar and Siddhartha Srinivasa.
\newblock A framework for push-grasping in clutter.
\newblock {\em Robotics: Science and systems VII}, 1, 2011.

\bibitem{yuan2018rearrangement}
Weihao Yuan, Johannes~A Stork, Danica Kragic, Michael~Y Wang, and Kaiyu Hang.
\newblock Rearrangement with nonprehensile manipulation using deep
  reinforcement learning.
\newblock In {\em 2018 IEEE International Conference on Robotics and Automation
  (ICRA)}, pages 270--277. IEEE, 2018.

\bibitem{hasan2020human}
Mohamed Hasan, Matthew Warburton, Wisdom~C Agboh, Mehmet~R Dogar, Matteo
  Leonetti, He~Wang, Faisal Mushtaq, Mark Mon-Williams, and Anthony~G Cohn.
\newblock Human-like planning for reaching in cluttered environments.
\newblock In {\em 2020 IEEE International Conference on Robotics and Automation
  (ICRA)}, pages 7784--7790. IEEE, 2020.

\bibitem{huang2020mechanical}
Huang Huang, Marcus Dominguez-Kuhne, Vishal Satish, Michael Danielczuk, Kate
  Sanders, Jeffrey Ichnowski, Andrew Lee, Anelia Angelova, Vincent Vanhoucke,
  and Ken Goldberg.
\newblock Mechanical search on shelves using lateral access x-ray.
\newblock In {\em 2021 IEEE/RSJ International Conference on Intelligent Robots
  and Systems (IROS)}, pages 2045--2052. IEEE, 2020.

\bibitem{lee2021tree}
Jinhwi Lee, Changjoo Nam, Jonghyeon Park, and Changhwan Kim.
\newblock Tree search-based task and motion planning with prehensile and
  non-prehensile manipulation for obstacle rearrangement in clutter.
\newblock In {\em 2021 IEEE International Conference on Robotics and Automation
  (ICRA)}, pages 8516--8522. IEEE, 2021.

\bibitem{zhong2022soft}
Sheng Zhong, Nima Fazeli, and Dmitry Berenson.
\newblock Soft tracking using contacts for cluttered objects to perform blind
  object retrieval.
\newblock {\em arXiv preprint arXiv:2201.10434}, 2022.

\bibitem{moll2017randomized}
Mark Moll, Lydia Kavraki, Jan Rosell, et~al.
\newblock Randomized physics-based motion planning for grasping in cluttered
  and uncertain environments.
\newblock {\em IEEE Robotics and Automation Letters}, 3(2):712--719, 2017.

\bibitem{jain2013reaching}
Advait Jain, Marc~D Killpack, Aaron Edsinger, and Charles~C Kemp.
\newblock Reaching in clutter with whole-arm tactile sensing.
\newblock {\em The International Journal of Robotics Research}, 32(4):458--482,
  2013.

\bibitem{ma2021extrinsic}
Daolin Ma, Siyuan Dong, and Alberto Rodriguez.
\newblock Extrinsic contact sensing with relative-motion tracking from
  distributed tactile measurements.
\newblock In {\em 2021 IEEE International Conference on Robotics and Automation
  (ICRA)}, pages 11262--11268. IEEE, 2021.

\bibitem{rezaei2021learning}
Sahand Rezaei-Shoshtari, Francois~Robert Hogan, Michael Jenkin, David Meger,
  and Gregory Dudek.
\newblock Learning intuitive physics with multimodal generative models.
\newblock {\em arXiv preprint arXiv:2101.04454}, 2021.

\bibitem{bhattacharjee2014inferring}
Tapomayukh Bhattacharjee, James~M Rehg, and Charles~C Kemp.
\newblock Inferring object properties from incidental contact with a
  tactile-sensing forearm.
\newblock {\em ArXiv}, 2014.

\bibitem{bhattacharjee2012haptic}
Tapomayukh Bhattacharjee, James~M Rehg, and Charles~C Kemp.
\newblock Haptic classification and recognition of objects using a tactile
  sensing forearm.
\newblock In {\em 2012 IEEE/RSJ International Conference on Intelligent Robots
  and Systems}, pages 4090--4097. IEEE, 2012.

\bibitem{pwlf}
Charles~F. Jekel and Gerhard Venter.
\newblock {\em {pwlf:} A Python Library for Fitting 1D Continuous Piecewise
  Linear Functions}, 2019.

\end{thebibliography}

\end{document}